\documentclass[letterpaper]{article} 
\usepackage{aaai25}  
\usepackage{times}  
\usepackage{helvet}  
\usepackage{courier}  
\usepackage[hyphens]{url}  
\usepackage{graphicx} 
\usepackage{natbib}  
\usepackage{caption} 
\usepackage{algorithm}
\usepackage{algorithmic}
\usepackage{amsmath,amsfonts}
\usepackage{tabularx}
\usepackage{multirow}
\usepackage{booktabs}
\usepackage{cite}
\usepackage{newfloat}
\usepackage{listings}
\DeclareCaptionStyle{ruled}{labelfont=normalfont,labelsep=colon,strut=off} 
\floatstyle{ruled}
\newfloat{listing}{tb}{lst}{}
\floatname{listing}{Listing}
\title{FD\textsuperscript{2}-Net: Frequency-Driven Feature Decomposition Network for Infrared-Visible Object Detection}
\author{
    Ke Li\textsuperscript{\rm 1},
    Di Wang\textsuperscript{\rm 1}\thanks{Corresponding author.},
    Zhangyuan Hu\textsuperscript{\rm 1},
    Shaofeng Li\textsuperscript{\rm 1}\footnotemark[1],
    Weiping Ni\textsuperscript{\rm 2},
    Lin Zhao\textsuperscript{\rm 3},
    Quan Wang\textsuperscript{\rm 1}
}
\affiliations{
    \textsuperscript{\rm 1}School of Computer Science and Technology, Xidian University\\
    \textsuperscript{\rm 2}Northwest Institute of Nuclear Technology\\
    \textsuperscript{\rm 3}School of Computer Science and Engineering, Nanjing University of Science and Technology\\
}

\usepackage{bibentry}

\begin{document}

\maketitle

\begin{abstract}
    Infrared-visible object detection (IVOD) seeks to harness the complementary information in infrared and visible images, thereby enhancing the performance of detectors in complex environments. 
    However, existing methods often neglect the frequency characteristics of complementary information, such as the abundant high-frequency details in visible images and the valuable low-frequency thermal information in infrared images, thus constraining detection performance.
    To solve this problem, we introduce a novel \textbf{F}requency-\textbf{D}riven \textbf{F}eature \textbf{D}ecomposition \textbf{Net}work for IVOD, called FD\textsuperscript{2}-Net, which effectively captures the unique frequency representations of complementary information across multimodal visual spaces.
    Specifically, we propose a feature decomposition encoder, wherein the high-frequency unit (HFU) utilizes discrete cosine transform to capture representative high-frequency features, while the low-frequency unit (LFU) employs dynamic receptive fields to model the multi-scale context of diverse objects. 
    Next, we adopt a parameter-free complementary strengths strategy to enhance multimodal features through seamless inter-frequency recoupling.
    Furthermore, we innovatively design a multimodal reconstruction mechanism that recovers image details lost during feature extraction, further leveraging the complementary information from infrared and visible images to enhance overall representational capacity.
    Extensive experiments demonstrate that FD\textsuperscript{2}-Net outperforms state-of-the-art (SOTA) models across various IVOD benchmarks, \textit{i.e.} LLVIP ($96.2\%$ mAP), FLIR ($82.9\%$ mAP), and M\textsuperscript{3}FD ($83.5\%$ mAP).
\end{abstract}

\section{Introduction}
\label{sec:intro}
Object detection is a foundational topic in computer vision, aiming to localize and identify diverse objects within images or videos.
It has extensive applications in autonomous driving, surveillance, and remote sensing \cite{10144688, li2023large}. 
Nevertheless, visible object detection encounters substantial challenges in adverse conditions like rain, fog, clouds, and poor illumination, primarily due to the inherent limitations of RGB sensors.
As a result, alternative visual sensors, particularly infrared cameras, are increasingly utilized to complement RGB cameras in overcoming these difficulties, thereby igniting substantial research interest in Infrared-Visible Object Detection (IVOD).

\begin{figure}[!t]
  \includegraphics[width=\linewidth]{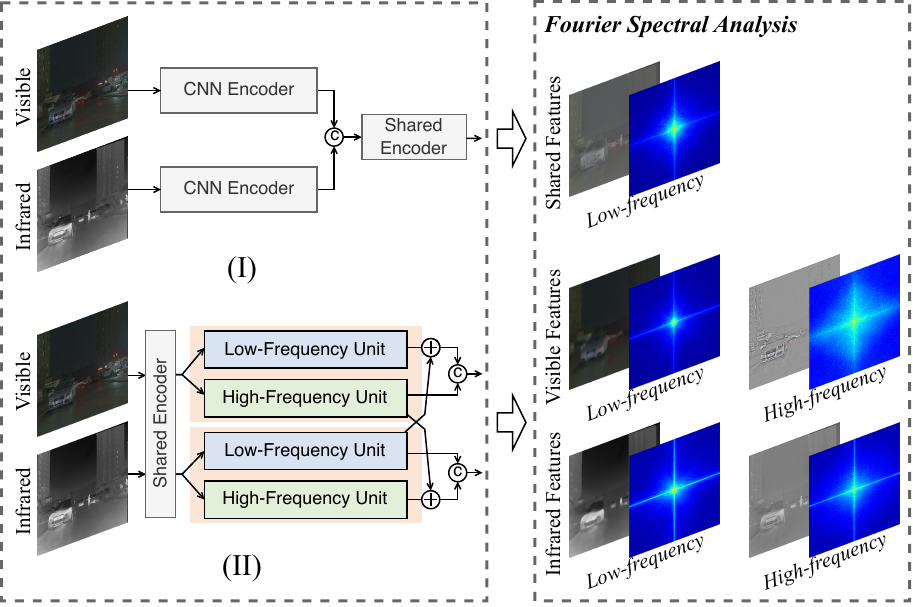}
  \caption{Illustration of the differences between our FD\textsuperscript{2}-Net and existing IVOD approaches.
  Our algorithm employs frequency decoupling to separate high- and low-frequency information in infrared and visible images, thereby effectively leveraging multimodal complementary features to extract more discriminative and robust characteristics.
  }
  \label{fig:teaser}
\end{figure}

However, current IVOD methods still have three weaknesses. 
\textit{\textbf{Weakness 1:}}
They tend to overlook the frequency characteristics of object features within infrared and visible images. 
Infrared imaging primarily captures low-frequency thermal radiation, while visible imaging emphasizes high-frequency details. 
Prevailing architectures \cite{li2023lrrnet, zhao2023ddfm} often overlook this kind of intrinsic property, and embed cross-modality information into a unified feature space, which results in the inability to extract modality-specific features.
\textit{\textbf{Weakness 2:}}
With a fixed receptive field, these methods only extract local information, which makes it difficult to adapt to the positional biases inherent in infrared and visible images.
Moreover, models with small kernels are inadequate for effectively capturing long-range information, which is crucial as the surrounding environment provides vital clues about object size, shape, and other characteristics \cite{Li_2024_CVPR}.
\textit{\textbf{Weakness 3:}} 
Recent IVOD approaches commonly employ downsampling operations to mitigate visual noise and reduce computational overhead, potentially resulting in the loss of object information.
Such degradation in feature representation significantly hampers the localization and classification capabilities of the detection head, ultimately compromising detection performance.
Our research explores a more rational paradigm to address these challenges in cross-modality feature extraction for IVOD tasks.
Based on the aforementioned analysis, we identify three critical countermeasures (CM):

\textit{\textbf{CM 1:}} 
We revisit the feature extraction process from a frequency perspective. 
Visible images furnish abundant high-frequency information, such as edges and textures, whereas infrared images deliver valuable low-frequency thermal radiation information.
As illustrated in Fig.~\ref{fig:teaser} (I), conventional methods depends solely on redundant cross-modality similar clues, leading to the loss of crucial complementary features.
In contrast, we can capture discriminative complementary information from infrared and visible images in a more controlled and interpretable manner by limiting the frequency space of feature extraction.
As shown in Fig.~\ref{fig:teaser} (II), adaptive frequency decoupling facilitates the retention of more representative low-frequency and high-frequency information in both infrared and visible images.

\textit{\textbf{CM 2:}} 
From a model design standpoint, larger kernels aid in capturing more extensive scene context, thereby mitigating geometric biases between infrared and visible images.
However, employing large kernel convolutions may introduce substantial background noise and overlook fine-grained details within the receptive field, which can be detrimental to the precise detection of small objects.
Hence, we parallelly arrange multiple depthwise dilated convolutions of varying sizes to extract multi-granularity texture features across diverse receptive fields, thus fulfilling IVOD tasks.

\textit{\textbf{CM 3:}} 
To combat the information loss resulting from repeated downsampling, many existing methods often employ generative approaches such as image super-resolution to alleviate this issue.
However, these methods not only require constructing pairs of high-resolution and low-resolution samples, but their generative processes often introduce spurious artifacts.
Conversely, we integrate a simple yet effective multimodal reconstruction mechanism into the IVOD framework, leveraging complementary information from both infrared and visible modalities to restore structural and texture details lost during feature extraction. 

In this paper, we design a novel paradigm for IVOD tasks, \textit{i.e.}, \textit{Frequency-Driven Feature Decomposition Network (\textbf{FD\textsuperscript{2}-Net})}, which decouples the frequency information of infrared and visible images to efficiently extract representative features and leverages the dominant frequency characteristics of one modality to enhance the complementary features of the other.
Specifically, we introduce a feature decomposition encoder, which comprises three main parts: a high-frequency unit (HFU), a low-frequency unit (LFU) and a parameter-free complementary strengths strategy (CSS).
HFU performs the discrete cosine transform, followed by a lightweight module that learns a spatial attention mask from multiple high-frequency components, thereby accentuating the most representative high-frequency features.
LFU employs multi-scale convolutional kernels to capture low-frequency structures of various objects and their contextual information, effectively modeling the relationships between objects and their surrounding environments.
Subsequently, CSS leverages the strengths of one modality to achieve complementary enhancement in the other.
Furthermore, we develop a cross-reconstruction unit (CRU) incorporating feature-level complementary masks.
CRU further learns complementary information from infrared and visible features through both fine-grained and coarse-grained cross-modality interactions, restoring the multimodal images.
Our contributions can be summarized as follows:

\begin{itemize}
  \item We propose a novel paradigm for IVOD, termed FD\textsuperscript{2}-Net, which aims to improve detection performance by effectively extracting valuable complementary features from infrared and visible images.
  \item We design a high-frequency unit (HFU) and a low-frequency unit (LFU) to effectively capture discriminative frequency information in both infrared and visible images. Also, a complementary strengths strategy is developed to enhance multimodal features through seamless inter-frequency recoupling.
  \item We introduce a cross-reconstruction unit (CRU) to integrate complementary information across modalities, thereby further enhancing feature representation.
  \item Extensive qualitative and quantitative experiments validate the effectiveness of our FD\textsuperscript{2}-Net, achieving accuracies of $96.2\%$ on LLVIP \cite{jia2021llvip}, $82.9\%$ on FLIR \cite{razakarivony2016vehicle}, and $83.5\%$ on M\textsuperscript{3}FD \cite{razakarivony2016vehicle}.
\end{itemize}

\section{Related work}
\label{sec2}
\subsection{General Object Detection} 
General object detectors can be broadly classified into two-stage detectors and one-stage detectors.
Faster R-CNN \cite{ren2015faster} is a classic two-stage detector, consisting of a Region Proposal Network (RPN), Region of Interest (RoI) pooling, and detection heads.
The RPN generates proposals based on features extracted by the backbone network. 
The extracted image features and generated proposals are fed into the RoI pooling operation to extract proposal features. 
Finally, the proposal features are classified and regressed by the detection head. 
To generate better region proposals, various methods have been explored to enhance performance, including architecture design \cite{cai2018cascade}, anchor box optimization \cite{jiang2018acquisition}, and multi-scale training \cite{singh2018sniper}.
However, two-stage methods necessitate filtering a large number of proposals, leading to significant time and computational overhead.
In contrast, one-stage detection frameworks predict bounding boxes and classes directly from densely sampled grids, thus achieving faster inference speeds.
YOLOv1 \cite{redmon2016you} is the first one-stage object detector to achieve real-time object detection. Through years of continuous development, the YOLO detectors have surpassed other one-stage object detectors \cite{liu2016ssd, lin2017focal} and become synonymous with real-time object detection. 
In this article, YOLO-based architicture is chosen as the detector to reasonably balance speed and accuracy.

\subsection{Infrared-Visible Object Detection} 
Infrared-visible fusion can complementarily capture richer object information, yielding more stable detection results.
The main focus of IVOD detectors has primarily been on exploring improved fusion techniques, for which several variant frameworks have been proposed.
TINet \cite{zhang2023illumination} enhances the extraction of complementary information by emphasizing the differences between infrared and visible images. 
AR-CNN \cite{Zhang_2019_ICCV} highlights that visible images and infrared images are misaligned in the spatial dimension. 
To align the regional features of two modalities, it proposes a region feature alignment module to enhance detection performance.
Furthermore, DMAF \cite{zhou2020improving} designs an illumination-aware feature alignment module that selects features based on illumination conditions and adaptively aligns features across modalities. 
To effectively capture the complementary features of infrared-visible images, APWNet \cite{zhang2023real} introduces an image fusion loss to enhance the performance of YOLOv5 \cite{Jocher_YOLOv5_by_Ultralytics_2020}. 
SuperYOLO \cite{zhang2023superyolo} adds an image super-resolution branch to strengthen the feature extraction capability of the backbone.
LRAF-Net \cite{10144688} improves detection performance by fusing the long-range dependencies of the visible and infrared features. 
DFANet \cite{10163477} introduces an antagonistic feature extraction and divergence module to extract the differential infrared and visible features with unique information.

In this paper, we propose a frequency-driven feature decomposition network that can efficiently extract discriminative complementary information from infrared and visible images, respectively.
This extracted information is then utilized to enhance feature representation, thereby improving detection performance.

\begin{figure*}[!t]
	\centerline{\includegraphics[width=\linewidth]{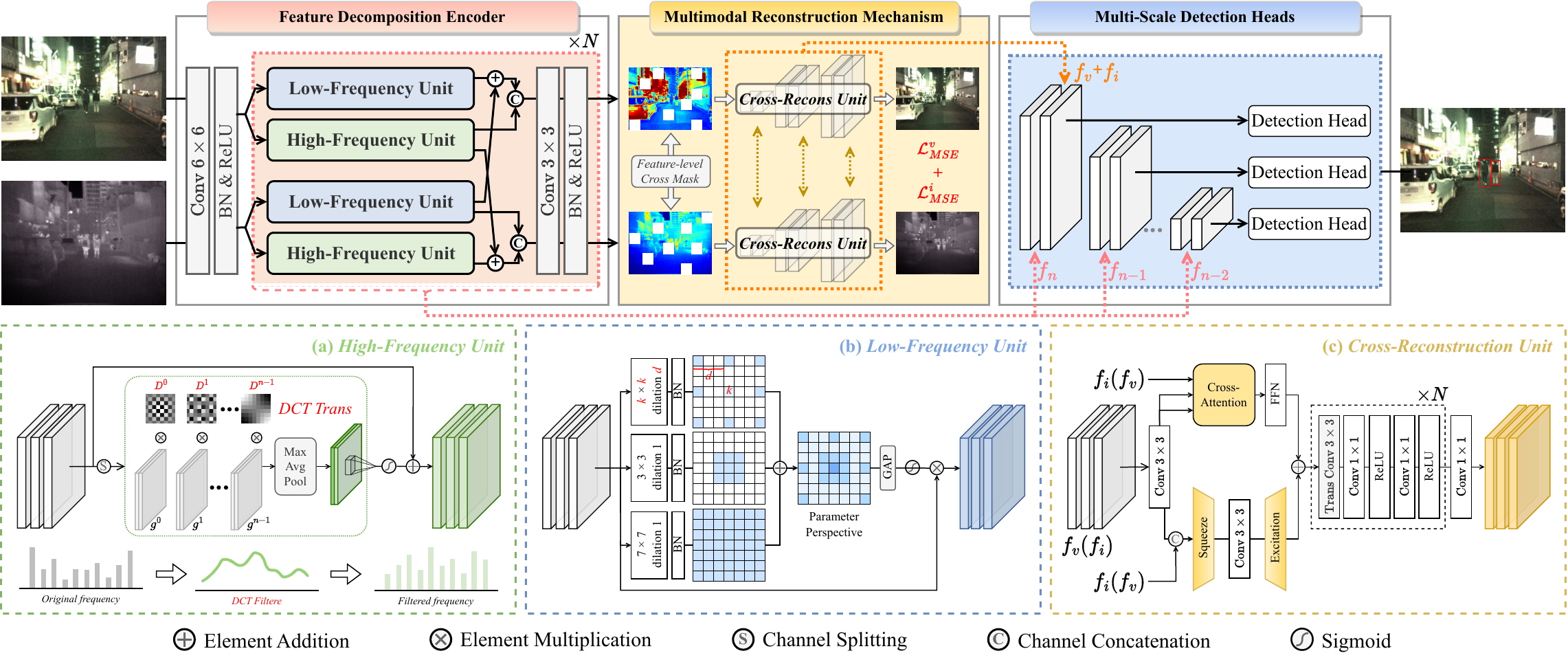}}
	\caption{The architecture (top row) and core components (bottom row) of our FD\textsuperscript{2}-Net.
  It has three components:
  (1) Feature Decomposition Encoder, which effectively extracts high/low-frequency features in multimodal visual space.
  (2) Multimodal Reconstruction Mechanism, which further learns the distinguishing and complementary features of each modality through the reconstruction of multimodal images to enhance feature representation. 
  (3) Multi-Scale Detection Head, which uses visual features from (1) and (2) to complete object classification and localization.}
	\label{Fig:workflow}
\end{figure*}

\section{Proposed Method}
\label{sec3}
\subsection{Overall Architecture}
\label{sec3-B}
As shown in Fig.~\ref{Fig:workflow}, our FD\textsuperscript{2}-Net comprises three modules:
\textbf{1) Feature Decomposition Encoder.}
Inspired by spectral spectrum, this module introduces a two-branch architecture to effectively extract valuable high-frequency and low-frequency features through feature decomposition and fusion.
Subsequently, through the complementary advantage strategy, the representative frequency features are reorganized to improve the overall representation ability.
\textbf{2) Multimodal Reconstruction Mechanism.}
To enhance feature learning, an asymmetric cross-maske strategy is applied to the features from the final layer of the Encoder, compelling each modality to obtain useful information from complementary modalities.
Two cross-reconstruction units are then used to restore the multimodal image by leveraging the complementary features of infrared and visible images.
The reconstruction process is constrained by the mean square error at the pixel level.
\textbf{3) Multi-Scale Detection Heads.}
This module constructs a Feature Pyramid Network (FPN) that utilizes multi-scale features extracted at various stages of the Encoder.
At the highest resolution layer of the FPN, the reconstructed multimodal features are integrated to further enhance detection.
Finally, following YOLOv5 \cite{Jocher_YOLOv5_by_Ultralytics_2020}, three detection heads with different scales are configured to accurately detect objects.

\subsection{Feature Decomposition Encoder}
Formally, let $I \in \mathbb{R}^{ H \times W}$ and $V \in \mathbb{R}^{3 \times H \times W}$ be the input infrared and visible images, where $H \times W$ represents the spatial resolution.
Initially, a $6\times6$ CBR block\footnote[1]{A $6\times6$ convolutional layer with a batch normalization (BN) \cite{ioffe2015batch} layer and a rectified linear unit (ReLU) \cite{nair2010rectified}.} is employed to reduce the resolution and extract shallow multimodal visual features $\{X_{I}^{S}, X_{V}^{S}\}\in \mathbb{R}^{c \times h \times w}$. 
Then, we first split $\{X_{I}^{S}, X_{V}^{S}\}$ into two components in a ratio of $\alpha$, respectively. 
One is expected to represent high-frequency component, denoted as $\Phi^H=\{X_{I}^{H},X_{V}^{H}\} \in \mathbb{R}^{\alpha c \times h \times w}$, capturing the spatial details such as edges and textures.
The other $\Phi^L=\{X_{I}^{L},X_{V}^{L}\} \in \mathbb{R}^{(1-\alpha) c \times h \times w}$ is expected to learn low-frequency content like context and structural information.

\subsubsection{High-Frequency Feature Attention.}
Our goal is to effectively extract high-frequency components from infrared and visible images, respectively. 
We thus introduce a High-Frequency Unit, which can filter out the high-frequency information and direct model's attention on more valuable information. 
The discrete cosine transform (DCT) has demonstrated superiority in image compression, particularly in enhancing image details and textures while eliminating noise.
Based on this, we incorporate DCT into IVOD.
This transformation guides the convolution to extract diverse high-frequency spatial features and effectively suppresses noises such as Gaussian and thermal noise in infrared-visible images.

\textbf{Discrete Cosine Transform (DCT).}
For an imgae $x \in \mathbb{R}^{H \times W}$, where $H$ and $W$ are the height and width of $x$, Eq.~(\ref{tab:equ21}) provides the definition of the standard two-dimensional (2D) DCT, mathematically defined as:
\begin{equation}
  \label{tab:equ21}
    f^{h,w} =  {\textstyle \sum_{i=0}^{H-1}}  {\textstyle \sum_{j=0}^{W-1}} x_{i,j} B^{h, w}_{i, j}, \\
\end{equation}
\begin{equation}
  \label{tab:equ22}
    B^{h, w}_{i, j} = \cos (\pi h/H  (i+1/2)) \cos (\pi w/W(j+1/2)), \\
\end{equation}
where $f \in \mathbb{R}^{H \times W}$ is the 2D DCT frequency spectrum, $B$ is the basis function of the 2D DCT, $h \in \{0,H-1\}$ and $w \in \{0,W-1\}$, and $\cos(\cdot)$ represents the cosine function.
To simplify the notation, constant normalization factors in Eq.~(\ref{tab:equ21}) are omitted.

\textbf{High-Frequency Unit.}
To adaptively modulate the emphasis on different frequency components for enhanced spatial information discrimination, we leverage the 2D DCT as a selective filtering mechanism.
Specifically, the high-frequency feature maps $\Phi^H$ are divided along the channel dimension into $n$ segments. 
Each group $\Phi^H_g$, where $g \in \{0,n-1\}$, maintains the spatial dimensions of $\Phi^H$ but has only $1/n$ of channel length. 
A specific 2D DCT frequency component, denoted as $B_{u_g,v_g}$, is then assigned to each segment and then concatenate to obatain the modality-specific high-frequency features, which is denoted as:
\begin{equation}
  \label{tab:equ4}
  \Phi^H = [\Phi^H_{0} * B_{u_0,v_0},\cdots,\Phi^H_{n-1} * B_{u_{n-1},v_{n-1}}],
\end{equation} 
where $[u_g,v_g]$ represents the 2D frequency indices corresponding to $\Phi^H_g$.
The $[\cdot, \cdot]$ is a concatenation operation.
Here, $g$ serves as the control parameter for frequency components, where a larger $g$ value enables channels within the same convolutional layer to capture multi-frequency features, thereby enhancing feature representation capability.

Next, we apply a spatial attention mechanism to adaptively learns a spatial mask to dynamically modulate different frequency components during training.
Mathematically, this instantiation can be formulated as:
\begin{equation}
  \label{tab:equ7}
  \text{SA}^{H} = \sigma(\mathcal{F}^{2\to1}_{7\times 7}[\text{AvgPool}(\Phi^H),\text{MaxPool}(\Phi^H)]),
\end{equation} 
where $\sigma$ denotes the sigmoid function. $\text{AvgPool}(\cdot)$ and $\text{MaxPool}(\cdot)$ are average-pooling and max-pooling operations.
$\mathcal{F}^{2\to1}$ is a $7\times7$ convolutional layer to transform the features (with 2 channels) into one spatial attention map, which facilitates information interaction among various spatial descriptors.

The final output of the HFU module is the element-wise product of the input feature $\Phi^H$ and $\text{SA}^{H}$, as indicated below:
\begin{equation}
  \label{tab:equ8}
  \Phi^{H'} = \text{SA}^{hf}\otimes \Phi^H.
\end{equation}
We believe using more complex attention architectures, such as \cite{behera2021context, bao2024relevant}, is of the potentials to achieve higher improvements.

\subsubsection{Low-Frequency Context Refinement.}
\label{sec3-B1}
To effectively capture low-frequency information across multiple scales, we construct a multi-granularity convolution by a set of parallel depth-wise convolutions (DWCs) with different kernel sizes and dilation rates.
For the $i$-th DWC, the expansion of the kernel size $k_i$ and dilation rate $d_i$ are flexible, with the only constraint being:
\begin{equation}
  \label{tab:equ9}
  k_i+(k_i-1)\times(d_i-1) \le RF.
\end{equation}
However such a multi-branch structure invariably increases computational cost, thereby prolonging inference times in practical deployments. 
References \cite{ding2019acnet,ding2021repvgg} point out that multiple parallel convolutional blocks can be seamlessly consolidated into a single convolutional layer for inference, optimizing computational efficiency.
By leveraging this equivalent transformation, we merge several small-kernel branches into a unified large-kernel convolutional layer, as shown in Fig.~\ref{Fig:workflow}.
This approach not only enhances the extraction of multi-scale features within a single layer but also maintains swift inference capabilities.
Following the ConvNeXt \cite{liu2022convnet} and RepLKNet \cite{ding2022scaling}, we set $RF = 7$, kernels size is $[7,3,3,3]$ and dilation rates is $[1,1,2,3]$.
Note that our LFU uses dilated convolution, thereby preventing the extraction of overly dense feature representations.

To diminish information redundancy and improve the feature diversity, we employ a channel-mix strategy that performs both inter-channel communications and spatial aggregations.
First, a Global Average Pooling (GAP) operation collates channel statistics from low-frequency spatial features.
These features then undergo compression and restoration via two sequential $1\times1$ convolution layers, reducing feature similarity.
A sigmoid function subsequently generates channel weights, refining the multi-scale spatial features $\Phi^L$ through weighted processing.
This process is encapsulated as follows:
\begin{equation}
  \label{tab:equ10}
  \Phi^{L'} = \Phi^L \otimes \sigma(\mathcal{F}^{d\to (1-\alpha) C}_{1\times1}(\mathcal{F}^{(1-\alpha) C\to d}_{1\times1}(\Phi^L))),
\end{equation}
where $d$ is set as $(1-\alpha) C/4$.

\subsubsection{Complementary Strengths Strategy.}
The function of this strategy is to recouple the complementary features and achieve efficient inter-frequency communication.
we propose a parameter-free manner to add the low/high-frequency features in cross-modality image to another features, where:
\begin{equation}
  \label{tab:equ111}
    X_I^{H'} \Leftarrow X_I^{H'}+X_V^{H'}, \\    
\end{equation}
\begin{equation}
  \label{tab:equ112}
    X_V^{L'} \Leftarrow X_V^{L'}+X_I^{L'}. \\    
\end{equation}
For each of frequency features witthin one modality, we concatenate both and use a $3\times3$ convolution layer $\mathcal{F}(\cdot)$ to obtain the enhanced modality-shared features.
The final output is formulated as follows:
\begin{equation}
  \label{tab:equ121}
    Y_{I}^{S} =\mathcal{F}([X_I^{H'},X_I^{H'}]), \\    
\end{equation}
\begin{equation}
  \label{tab:equ122}
    Y_{V}^{S} =\mathcal{F}([X_V^{H'},X_V^{H'}]). \\    
\end{equation}
Fig.~\ref{Fig:workflow} shows a detailed illustration of the LFU, HFU and fusion stragy, where we intuitively demonstrate how they work by synergistically capturing high/low-frequency spatial information.

\subsection{Multimodal Reconstruction Mechanism}
As mentioned above, the Feature Decomposition Encoder focuses on explicitly extracting valuable frequency information.
To take full advantage of complementary information, we further integrate a Multimodal Reconstruction Mechanism into our FD\textsuperscript{2}-Net.
It aims to learn the discriminative and complementary features of each modality while augmenting the overall representation capability.
As showcased in Fig.~\ref{Fig:workflow}, this mechanism has two components: feature-level cross-maske and Cross-Reconstruction Unit (CRU).

\subsubsection{Feature-Level Complementary Mask.}
To better utilize the multimodal information, avoid the network always learning from a single image.
We design an efficient feature augmentation strategy to train FD\textsuperscript{2}-Net.
As shown in Fig.~\ref{Fig:workflow}, we perform an asymmetric mask of local information, which denoted as:
\begin{equation}
  \label{tab:equ13}
    \text{M}_{all} = \text{M}_I \bigcup \text{M}_V,\ \text{M}_I \mid \text{M}_V = 1,
\end{equation}
where $\text{M}_I$ and $\text{M}_V$ represent the infrared mask and visible mask, respectively.
$\text{M}_{all}$ represents the total unseen area, accounting for 30\% of the feature map.
Such a design allows the network to only obtain valid information from the position corresponding to the opposite modality of the masked area. 

\subsubsection{Multimodal Image Reconstruction.}
As mentioned in the introduction, the information loss caused by feature extraction leads to difficulties for the detector to localize and identify objects.
To address the challenge, we introduce Cross-Reconstruction Unit (CRU) to learn the complementary features through fine-grained local and coarse-grained global interactions.
Note that CRU is a generic image reconstruction network, and we only take the visible image as an example to explain the working of CRU. 
The process can be expressed as follows (where the Rectified Linear Unit (ReLU) is omitted for brevity):
\begin{equation}
  \label{tab:equ141}
    x_v = Conv_{3\times3}(x_v),\\
\end{equation}
\begin{equation}
  \label{tab:equ142}
    x_v'  = \text{CA}(x_v,x_i) + \mathcal{F}_E(Conv_{3\times3}(\mathcal{F}_S([x_v,x_i]))), \\
\end{equation}
\begin{equation}
  \label{tab:equ143}
    f_v = Conv_{1\times1}(Conv_{1\times1}(TransConv(x_v'))), \\
\end{equation}
where $\text{CA}(\cdot)$ represents the cross-attention layer.
$\mathcal{F}_S$ and $\mathcal{F}_E$ are feature squeeze and excitation operations, same as \cite{hu2018squeeze}.
For the infrared and visible image, the outputs of CRU are $f_i$ and $f_v$.

\subsection{Training Loss}
The total loss function comprises the image reconstruction loss $\mathcal{L}_{rc}$ and the detection loss $\mathcal{L}_{det}$.
The reconstruction loss is computed using the mean squared error (MSE) loss between the original and reconstructed images, which is formulated as follows:
\begin{equation}
  \label{tab:equ15}
  \mathcal{L}_{rc} = 1/2\left \| f_i - I  \right \| _{2} + 1/2\left \| f_v - V  \right \| _{2},
\end{equation}
where $f_i$ and $f_v$ are the reconstructed infrared and visible features, respectively.
$I$ and $V$ denote the input infrared and visible images, respectively.
The detection loss, consistent with the previous algorithm, comprises classification loss $\mathcal{L}_{cls}$, localization loss $\mathcal{L}_{box}$, and confidence loss $\mathcal{L}_{obj}$:
\begin{equation}
  \label{tab:equ16}
  \mathcal{L}_{det} = \mathcal{L}_{cls} + \mathcal{L}_{box} + \mathcal{L}_{obj}.
\end{equation}
The overall loss function is defined as follows:
\begin{equation}
  \label{tab:equ17}
  \mathcal{L}_{total} = \lambda_1\mathcal{L}_{rc} + \lambda_2\mathcal{L}_{det}.
\end{equation}
The $\lambda_1$ and $\lambda_2$ are the hyperparameters to balance the two losses during training.

\section{Experiments}
\label{sec4}
\subsection{Experimental Settings}
\subsubsection{Datasets.}
\label{sec4a}
The proposed model is evaluated against SOTA methods using three IVOD benchmark datasets:
(1) \textbf{LLVIP} dataset \cite{jia2021llvip} is a prominent large-scale pedestrian dataset specifically collected in low-light conditions, predominantly showcasing extremely dark scenes.
It ensures meticulous spatial and temporal alignment between all infrared and visible image pairs, concentrating solely on pedestrian detection.
(2) \textbf{FLIR} dataset offers a highly challenging multispectral object detection benchmark, encompassing both day and night scenes.
In this study, we utilized the “aligned” version \cite{zhang2020multispectral}.
It comprises 5,142 precisely aligned infrared-visible image pairs, with 4,129 pairs allocated for training and 1,013 pairs reserved for testing.
The dataset encompasses three primary object categories: People, Cars, and Bicycles.
(3) \textbf{M\textsuperscript{3}FD} dataset \cite{liu2022target} comprises 4,200 pairs of RGB and thermal images. 
It includes six categories of objects: People, Cars, Buses, Motorcycles, Lamps, and Trucks.
Following prior work \cite{zhao2023cddfuse}, we employ a random splitting method to delineate the training and validation sets.
Specifically, 80\% of the images are allocated to the training set, with the remaining images assigned to the validation set.

\begin{table}[!t]
	\centering
	\resizebox{\linewidth}{!}{
  \begin{tabular}{cccccc}
    \toprule
    Methods         & $\uparrow$F1   & $\uparrow$Precision & $\uparrow$Recall  & $\uparrow$$mAP_{50}$ & $\uparrow$$mAP_{75}$\\
    \hline
    Infrared        & 82.6 & 89.8 & 76.5     & 87.6  & 45.9\\
    Visible         & 83.5 & 90.7 & 77.4     & 88.5  & 46.3\\
    DensFuse        & 88.8 & 91.5 & 86.4     & 89.4  & 58.2\\
    SDNet           & 88.8 & 90.5 & 87.2     & 90.8  & 63.1\\
    U2Fusion        & 89.4 & 90.5 & 88.3     & 91.2  & 61.5\\
    CDDFuse         & 90.9 & 90.5 & \textbf{91.3}     & 93.6  & 65.7\\    
    MetaF           & 88.6 & 91.1 & 86.3     & 92.7  & 65.5\\
    LRRNet          & 91.4 & 93.1 & 89.9     & 94.8  & 68.8\\
    SegMiF          & 91.3 & \underline{93.5} & 89.2     & 94.3  & 67.1\\
    TarDAL          & 89.9 & 92.3 & 87.6     & 93.3  & 62.4\\
    DDFM            & 90.9 & 93.0 & 88.9     & 94.1  & 64.6\\
    CSSA            & 89.3 & 91.6 & 87.5 & 92.7 & 65.3 \\
    TFDet           & \underline{91.5} & 92.5 & \underline{90.4} & \underline{95.4} & \underline{68.9} \\
    \hline
    \textbf{Ours}   & \textbf{91.7} & \textbf{94.2} & 89.4     & \textbf{96.2}  & \textbf{70.0}\\
    \bottomrule		
    \end{tabular}}
    \caption{Comparison of FD\textsuperscript{2}Net and SOTA methods on \textbf{LLVIP} dataset. The best and second best performance are highlighted in \textbf{bold} and \underline{underline}.}
	\label{tab:LLVIP}
\end{table}

\subsubsection{Implementation Details.}
\label{sec4b}
To ensure fairness, we follow the same dataset processing approach as other mainstream methods \cite{fu2023lraf}.
FD\textsuperscript{2}Net is built upon the SOTA detector YOLOv5 \cite{Jocher_YOLOv5_by_Ultralytics_2020}. 
For evaluation, we report F1-Score, Precision, Recall, and Average Precision, consistent with prior research.
Xavier initialization \cite{glorot2010understanding} is used to initialize parameters, and the model is trained for 150 epochs using SGD \cite{robbins1951stochastic} with an initial learning rate of $0.01$, weight decay of $10^{-4}$, and momentum of $0.9$.

\subsection{Main Results}
\label{sec4e}
We compare our proposed FD\textsuperscript{2}Net with several baseline and SOTA methods, including SDNet \cite{zhang2021sdnet}, TarDAL \cite{liu2022target}, DensFuse \cite{li2018densefuse}, U2Fusion \cite{xu2020u2fusion}, CDDFuse \cite{zhao2023cddfuse}, SegMiF \cite{liu2023multi}, DDFM \cite{zhao2023ddfm}, MetaF \cite{zhao2023metafusion}, LRRNet \cite{li2023lrrnet}, CSSA \cite{cao2023multimodal}, and TFDet \cite{zhang2024tfdet}. 
These methods are built on the YOLOv5 detector to measure their detection performance.

\textbf{Comparison Results on LLVIP.}
The results presented in Table \ref{tab:LLVIP} demonstrate that our method effectively fuses similar and complementary features in infrared and visible images, significantly enhancing the network's representational capability. 
Compared to single-modality methods, FD\textsuperscript{2}Net outperforms both Infrared and Visible, with substantial improvements of \textbf{8.6\%} and \textbf{7.7\%}, respectively. 
Furthermore, when compared to other SOTA networks, FD\textsuperscript{2}Net consistently surpasses them, showing an improvement in $mAP_{50}$ by \textbf{1.4\%-6.8\%}. 
These results indicate that our proposed method markedly enhances IVOD tasks performance.

\textbf{Comparison Results on FLIR.} 
As illustrated in Table \ref{tab:FLIR}, FD\textsuperscript{2}Net demonstrates exceptional performance, establishing new SOTA benchmarks for $mAP_{50}$ and $mAP_{75}$ at \textbf{82.9\%} and \textbf{41.9\%}, respectively. 
Specifically, our method surpasses CDDFuse and SegMiF by \textbf{+2.1\%} and \textbf{+1.4\%} in terms of $mAP_{50}$. 
When the threshold is increased to 0.75, the miss rate for other methods rises more significantly than for FD\textsuperscript{2}Net, indicating our method's superior detection accuracy. 
For instance, it achieves \textbf{42.5\%} $mAP_{75}$, improving by \textbf{1.5\%} over the previous best model, LRRNet.

\begin{table}[!t]
	\centering
	\resizebox{\linewidth}{!}{
	\begin{tabular}{cccccc}
    \toprule
    Methods       & People  & Car   & Bicycle   & $\uparrow$$mAP_{50}$ & $\uparrow$$mAP_{75}$\\
    \hline
    Infrared      & 77.2 & 85.2 & 57.9 & 73.7 & 34.0  \\
    Visible       & 65.6 & 73.8 & 48.7 & 62.7 & 25.9  \\
    DensFuse      & 78.7 & 85.8 & 61.4 & 75.3 & 35.0 \\
    SDNet         & 81.0 & 87.3 & 64.2 & 77.5 & 33.1 \\
    U2Fusion      & 82.7 & 87.8 & 67.7 & 79.4 & 36.5 \\
    CDDFuse       & 82.3 & 87.2 & \underline{72.9} & 80.8 & 39.4 \\
    MetaF         & 83.3 & \underline{89.2} & 71.1 & 81.4 & 40.7 \\
    LRRNet        & 83.3 & 88.8 & 69.7 & 80.6 & 41.0 \\
    SegMiF        & \textbf{85.3} & 86.9 & 72.8 & 81.5 & 40.9 \\
    TarDAL        & 85.1 & 85.3 & 69.3 & 79.9 & 37.9 \\
    DDFM          & 84.5 & 87.9 & 71.5 & 81.2 & 40.2 \\
    CSSA          & 83.2 & 86.7 & 68.6 & 79.4 & 37.2 \\
    TFDet         & \underline{85.2} & 87.5 & 71.9 & \underline{81.7} & \underline{41.3} \\
    \hline
    \textbf{Ours} & \textbf{85.3} & \textbf{89.9} & \textbf{73.2} & \textbf{82.9} & \textbf{42.5} \\
    \bottomrule
	\end{tabular}}
    \caption{Comparison of FD\textsuperscript{2}Net and SOTA methods on \textbf{FLIR} dataset. The best and second best performance are highlighted in \textbf{bold} and \underline{underline}.}
	\label{tab:FLIR}
\end{table}
\begin{table}[!t]
	\centering
	\resizebox{\linewidth}{!}{
	\begin{tabular}{cccccccc}
    \toprule
    Methods       & Peo  & Car  & Bus  & Mot  & Lam  & Tru  & $\uparrow$$mAP_{50}$  \\
    \hline
    Infrared      & 80.6 & 88.7 & 78.6 & 63.7 & 69.9 & 66.2 & 74.6  \\
    Visible       & 69.4 & 90.6 & 78.7 & 69.3 & 86.2 & 71.4 & 77.6  \\
    DensFuse      & 76.3 & 91.8 & 79.3 & 72.7 & 77.0 & 72.5 & 78.4  \\
    SDNet         & 79.5 & 92.6 & 81.0 & 67.1 & 84.2 & 69.4 & 79.0  \\
    U2Fusion      & 77.3 & 91.3 & 81.1 & 73.0 & 85.1 & \underline{72.8} & 80.1  \\
    CDDFuse       & 81.1 & 93.2 & \underline{82.3} & 74.0 & 87.7 & 72.7 & 81.9  \\
    MetaF         & 81.6 & 93.3 & 81.9 & 74.8 & 87.3 & 70.8 & 81.6 \\
    LRRNet        & 79.7 & 92.0 & 80.4 & 73.6 & 86.5 & 68.8 & 80.2 \\
    SegMiF        & \underline{82.4} & \underline{93.4} & 81.8 & \underline{75.7} & 86.7 & 71.1 & \underline{82.2}  \\
    TarDAL        & 81.0 & 93.2 & 81.5 & 71.2 & 87.0 & 68.2 & 80.6  \\
    DDFM          & 82.0 & 93.1 & 82.2 & 73.6 & \textbf{87.9} & 71.0 & 81.7  \\
    \hline
    \textbf{Ours} & \textbf{83.7} & \textbf{93.6} & \textbf{82.7} & \textbf{78.1} & \underline{87.8} & \textbf{73.8} & \textbf{83.5}  \\
    \bottomrule
	\end{tabular}}
    \caption{Comparison of FD\textsuperscript{2}Net and SOTA methods on \textbf{M\textsuperscript{3}FD} dataset. The best and second best performance are highlighted in \textbf{bold} and \underline{underline}.}
	\label{tab:M3FD}
\end{table}

\textbf{Comparison Results on M\textsuperscript{3}FD.}
The comparative results on the M\textsuperscript{3}FD dataset are summarized in Table \ref{tab:M3FD}. 
Our proposed method achieves a $mAP_{50}$ of \textbf{83.5\%}, establishing new records. 
In addition, we present the detection accuracy for each category. 
Notably, in the “People” and “Motorcycle” categories, FD\textsuperscript{2}Net achieves improvements of \textbf{1.3\%} and \textbf{2.4\%} over the previous best method. 
This suggests that our method possesses a superior ability to detect weak and small objects.


\textbf{Visual Comparisons.}
The qualitative results are depicted in Fig.~\ref{fig:featuremap}. 
The green boxes are detection results, while the red dashed boxes mark missed objects (false negatives).
It is evident that the predictions made by previous methods suffer from missed detections, especially for small and occluded objects in images.
FD\textsuperscript{2}Net effectively captures robust shared and discriminative specific information related to detected objects, resulting in superior performance across various challenging scenarios.

\begin{figure}[!t]
	\centerline{\includegraphics[width=\linewidth]{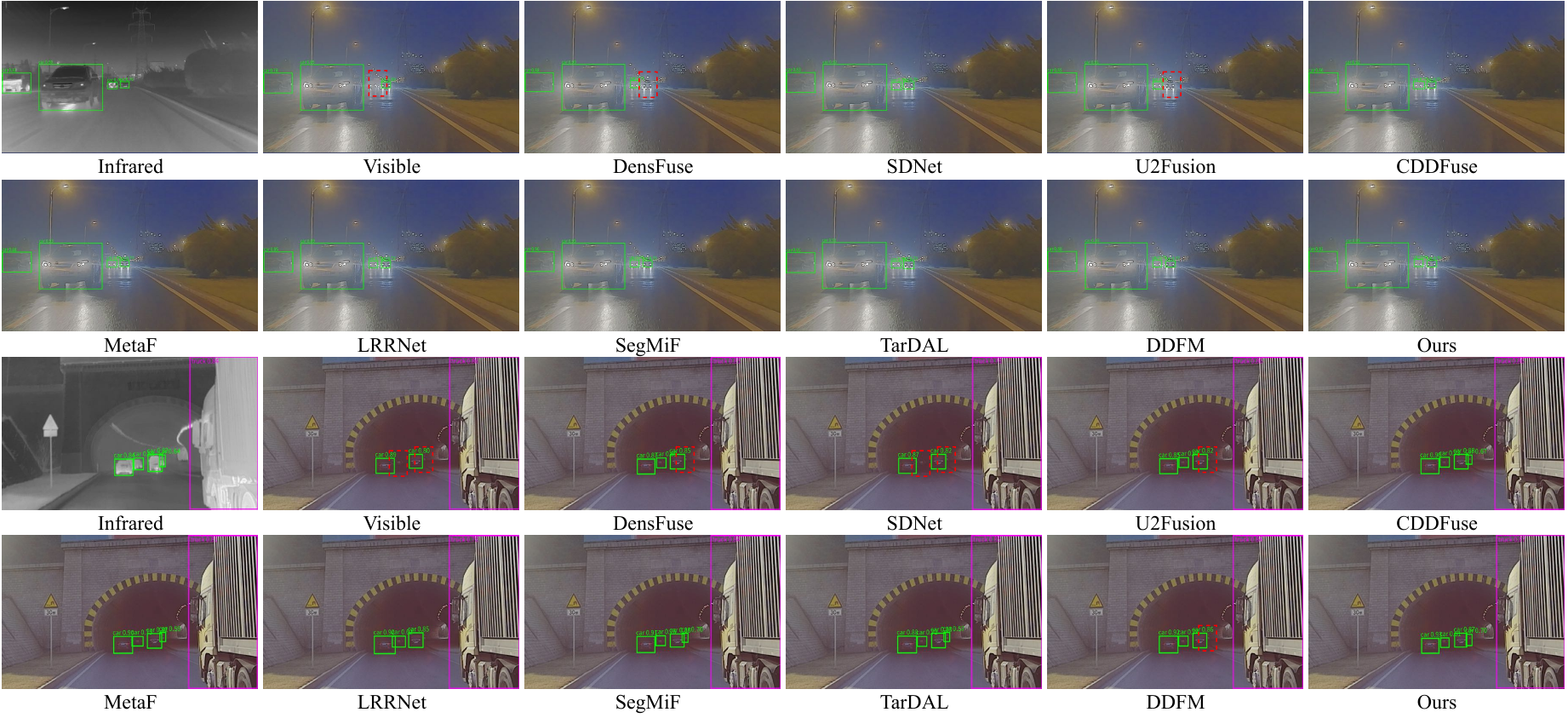}}
	\caption{Visual comparison of FD\textsuperscript{2}Net with 10 SOTA methods. Green boxes are detection results, while red dashed boxes mark missed objects (false negatives).}
	\label{fig:featuremap}
\end{figure}

\begin{table}[!t]
  \centering
  \resizebox{\linewidth}{!}{
  \begin{tabular}{l|cccc|c|c}
    \toprule
         &\multicolumn{2}{c}{Freq-Dec Encoder} & \multicolumn{2}{c|}{Cross-Rec Module}  & \multirow{2}{*}{$\uparrow$$mAP_{50}$} & \multirow{2}{*}{$\uparrow$$mAP_{75}$} \\
    \cline{2-5}
         & HFU        & LFU        &  CRU       &  Mask        & \\
    \hline 
    I    & -         & -          & -          & -          & 90.9 & 62.7 \\
    II   &\checkmark & -          & -          & -          & 92.4 & 64.4 \\
    III  &\checkmark & \checkmark & -          & -          & 94.7 & 66.3 \\
    IV   &\checkmark & \checkmark & \checkmark & -          & 95.6 & 69.6 \\
    \hline
    \textbf{Ours} &\checkmark & \checkmark & \checkmark & \checkmark & \textbf{96.2} & \textbf{70.0} \\
    \bottomrule
  \end{tabular}}
  \caption{Ablation study of FD\textsuperscript{2}Net components. HFU: High-Frequency Unit, LFU: Low-Frequency Unit, CRU: Cross-Reconstruction Unit. Mask: Complementary Mask Strategy.}
  \label{tab:ablation}
\end{table}

\subsection{Ablation Study}
\label{sec4c}
In this section, we present the ablation study results on the LLVIP dataset to evaluate the relative effectiveness of different components in FD\textsuperscript{2}Net.

\textbf{Architecture of FD\textsuperscript{2}Net.}
Compared to the baseline (Exp.I), the introduction of HFU (Exp.II) and LFU (Exp.III) for enhanced feature extraction improves $mAP_{50}$ by \textbf{1.5\%} and \textbf{2.3\%}, respectively.
Incorporating the multimodal image reconstruction strategy CRU (Exp.IV) into FD\textsuperscript{2}Net results in a \textbf{0.9\%} improvement in $mAP_{50}$.
Notably, $mAP_{75}$ exhibits a substantial improvement of \textbf{3.3\%}, indicating that object position perception can be significantly enhanced through image reconstruction.
The feature representation capability can be further enhanced by employing an asymmetric feature mask, leading to increases of \textbf{2.6\%} in $AP_{50}$ and \textbf{1.8\%} in $AP_{75}$.
These ablation results show the effectiveness of the major components in the proposed method.

\textbf{Effect of HFU and LFU.}
Our Feature Decomposition Encoder (FDE) comprises two components: high-frequency attention (HFU) and low-frequency refinement (LFU). 
To evaluate their effectiveness, we replaced the C2f blocks in YOLOv5n with either HFU or LFU.
As shown in Table~\ref{tab:combination}, using HFU (YOLOv5n+H) or LFU (YOLOv5n+L) alone resulted in performance drops of 3.2\% and 2.8\%, respectively, indicating that neither component alone effectively captures the complementary features of infrared-visible images.
We further explored three integration strategies: sequential high-to-low (H+L), sequential low-to-high (L+H), and parallel (H\&L). 
The parallel combination achieved the best performance, significantly improving $mAP_{50}$ of \textbf{94.7\%}, with reduced parameters and FLOPs. 
Thus, we adopt the parallel (H\&L) design for FDE to maximize model performance.

\textbf{Feature maps visualization.}
To investigate the feature representation capabilities of the proposed FD\textsuperscript{2}Net, we visualize the feature maps from the second stage of both the original YOLOv5 and FD\textsuperscript{2}Net.
As illustrated in Fig.~\ref{fig:featuremap}, the feature patterns produced by FD\textsuperscript{2}Net are significantly enriched compared to the original YOLOv5.
This approach not only reduces redundant features but also strengthens and diversifies representative features.


\begin{table}[!t]
  \centering
  \resizebox{\linewidth}{!}{
  \begin{tabular}{l|l|c|c|c}
      \toprule
      &Description                          & $\downarrow$Params.        & $\downarrow$FLOPs          & $\uparrow$$mAP_{50}$       \\
      \hline
      I&YOLOv5n                             &  3.01 M          & 8.1 G          & 90.9 \%          \\
      II&YOLOv5n + H                        &  \textbf{2.72 M} & \textbf{7.6 G} & 91.5 \%          \\
      III&YOLOv5n + L                       &  2.73 M          & 7.7 G          & 91.9 \%          \\
      IV&YOLOv5n + H + L                    &  2.77 M          & 8.0 G          & 92.8 \%          \\
      V&YOLOv5n + L + H                     &  2.77 M          & 8.0 G          & 92.6 \%          \\
      VI&YOLOv5n + H \& L                   &  2.75 M          & 7.8 G          & \textbf{94.7} \% \\
      \bottomrule		
  \end{tabular}}
  \caption{Expermental results with different combination methods of LFU and HFU on LLVIP dataset.}
  \label{tab:combination}
\end{table}
\begin{figure}[!t]
	\centerline{\includegraphics[width=\linewidth]{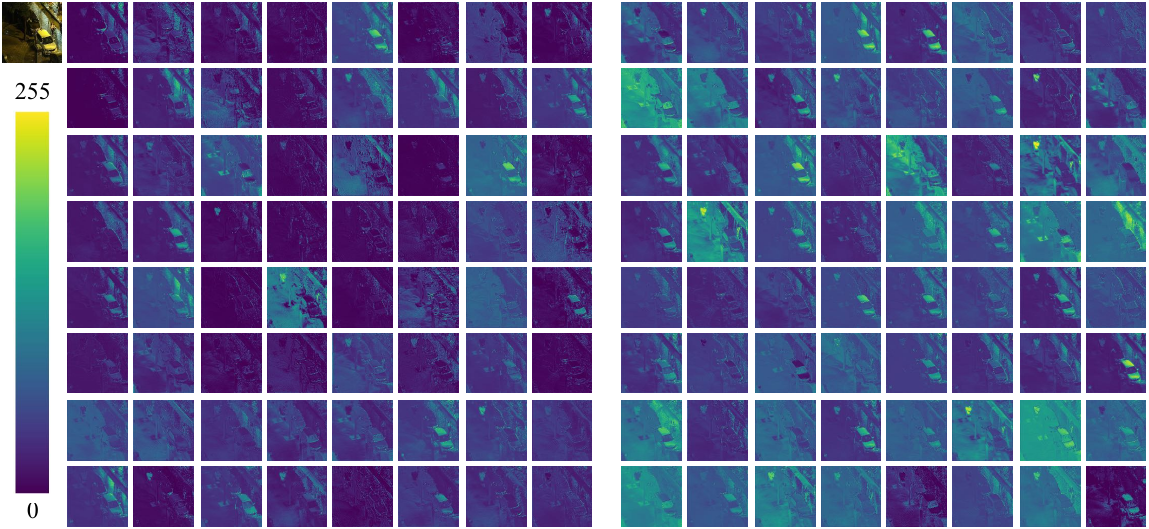}}
	\caption{Left: Features from the original YOLOv5n, Right: Features from the proposed FD\textsuperscript{2}Net.}
	\label{fig:featuremap}
\end{figure}


\section{Conclusion}
\label{sec5}
In this paper, we introduce a Frequency-Driven Feature Decomposition Network (FD\textsuperscript{2}Net) specifically designed for infrared-visible object detection tasks.
It efficiently models high-frequency and low-frequency features, thereby facilitating the extraction of valuable complementary information.
Furthermore, aided by the multimodal reconstruction mechanism, the complementary information within multimodal images is more effectively exploited.
Extensive qualitative and quantitative experiments demonstrate that the proposed network attains state-of-the-art performance across competitive infrared-visible object detection benchmarks.

\section{Acknowledgments}
This work was supported in part by the National Science and Technology Major Project under Grant 2022ZD0117103,  in part by the National Natural Science Foundation of China under Grants 62072354 and 62172222,  in part by the Fundamental Research Funds for the Central Universities under Grants QTZX23084 and XJSJ24015,  in part by the Natural Science Basic Research Program of Shaanxi under Grants 2024JC-YBQN-0732 and 2024JC-YBQN-0340, and in part by a grant from the Innovation Capability Support Program of Shaanxi under Grant 2023-CX-TD-08.

\bibliography{aaai25}

\begin{thebibliography}{43}
\providecommand{\natexlab}[1]{#1}

\bibitem[{Bao et~al.(2024)Bao, Qin, Sun, Wang, and Zheng}]{bao2024relevant}
Bao, X.; Qin, J.; Sun, S.; Wang, X.; and Zheng, Y. 2024.
\newblock Relevant Intrinsic Feature Enhancement Network for Few-Shot Semantic
  Segmentation.
\newblock In \emph{Proceedings of the AAAI Conference on Artificial
  Intelligence}, volume~38, 765--773.

\bibitem[{Behera et~al.(2021)Behera, Wharton, Hewage, and
  Bera}]{behera2021context}
Behera, A.; Wharton, Z.; Hewage, P.~R.; and Bera, A. 2021.
\newblock Context-aware attentional pooling (cap) for fine-grained visual
  classification.
\newblock In \emph{Proceedings of the AAAI conference on artificial
  intelligence}, volume~35, 929--937.

\bibitem[{Cai and Vasconcelos(2018)}]{cai2018cascade}
Cai, Z.; and Vasconcelos, N. 2018.
\newblock Cascade r-cnn: Delving into high quality object detection.
\newblock In \emph{Proceedings of the IEEE Conference on Computer Vision and
  Pattern Recognition}, 6154--6162.

\bibitem[{Cao et~al.(2023)Cao, Bin, Hamari, Blasch, and
  Liu}]{cao2023multimodal}
Cao, Y.; Bin, J.; Hamari, J.; Blasch, E.; and Liu, Z. 2023.
\newblock Multimodal object detection by channel switching and spatial
  attention.
\newblock In \emph{Proceedings of the IEEE/CVF Conference on Computer Vision
  and Pattern Recognition}, 403--411.

\bibitem[{Ding et~al.(2019)Ding, Guo, Ding, and Han}]{ding2019acnet}
Ding, X.; Guo, Y.; Ding, G.; and Han, J. 2019.
\newblock Acnet: Strengthening the kernel skeletons for powerful cnn via
  asymmetric convolution blocks.
\newblock In \emph{Proceedings of the IEEE/CVF International Conference on
  Computer Vision}, 1911--1920.

\bibitem[{Ding et~al.(2022)Ding, Zhang, Han, and Ding}]{ding2022scaling}
Ding, X.; Zhang, X.; Han, J.; and Ding, G. 2022.
\newblock Scaling up your kernels to 31x31: Revisiting large kernel design in
  cnns.
\newblock In \emph{Proceedings of the IEEE/CVF Conference on Computer Vision
  and Pattern Recognition}, 11963--11975.

\bibitem[{Ding et~al.(2021)Ding, Zhang, Ma, Han, Ding, and
  Sun}]{ding2021repvgg}
Ding, X.; Zhang, X.; Ma, N.; Han, J.; Ding, G.; and Sun, J. 2021.
\newblock Repvgg: Making vgg-style convnets great again.
\newblock In \emph{Proceedings of the IEEE/CVF Conference on Computer Vision
  and Pattern Recognition}, 13733--13742.

\bibitem[{Fu et~al.(2023{\natexlab{a}})Fu, Wang, Duan, Xiao, Dian, Li, and
  Li}]{fu2023lraf}
Fu, H.; Wang, S.; Duan, P.; Xiao, C.; Dian, R.; Li, S.; and Li, Z.
  2023{\natexlab{a}}.
\newblock Lraf-net: Long-range attention fusion network for visible--infrared
  object detection.
\newblock \emph{IEEE Transactions on Neural Networks and Learning Systems}.

\bibitem[{Fu et~al.(2023{\natexlab{b}})Fu, Wang, Duan, Xiao, Dian, Li, and
  Li}]{10144688}
Fu, H.; Wang, S.; Duan, P.; Xiao, C.; Dian, R.; Li, S.; and Li, Z.
  2023{\natexlab{b}}.
\newblock LRAF-Net: Long-Range Attention Fusion Network for Visible–Infrared
  Object Detection.
\newblock \emph{IEEE Transactions on Neural Networks and Learning Systems},
  1--14.

\bibitem[{Glorot and Bengio(2010)}]{glorot2010understanding}
Glorot, X.; and Bengio, Y. 2010.
\newblock Understanding the difficulty of training deep feedforward neural
  networks.
\newblock In \emph{Proceedings of the International Conference on Artificial
  Intelligence and Statistics}, 249--256.

\bibitem[{Hu, Shen, and Sun(2018)}]{hu2018squeeze}
Hu, J.; Shen, L.; and Sun, G. 2018.
\newblock Squeeze-and-excitation networks.
\newblock In \emph{Proceedings of the IEEE Conference on Computer Vision and
  Pattern Recognition}, 7132--7141.

\bibitem[{Ioffe and Szegedy(2015)}]{ioffe2015batch}
Ioffe, S.; and Szegedy, C. 2015.
\newblock Batch normalization: Accelerating deep network training by reducing
  internal covariate shift.
\newblock In \emph{International Conference on Machine Learning}, 448--456.

\bibitem[{Jia et~al.(2021)Jia, Zhu, Li, Tang, and Zhou}]{jia2021llvip}
Jia, X.; Zhu, C.; Li, M.; Tang, W.; and Zhou, W. 2021.
\newblock LLVIP: A visible-infrared paired dataset for low-light vision.
\newblock In \emph{Proceedings of the IEEE/CVF International Conference on
  Computer Vision}, 3496--3504.

\bibitem[{Jiang et~al.(2018)Jiang, Luo, Mao, Xiao, and
  Jiang}]{jiang2018acquisition}
Jiang, B.; Luo, R.; Mao, J.; Xiao, T.; and Jiang, Y. 2018.
\newblock Acquisition of localization confidence for accurate object detection.
\newblock In \emph{Proceedings of the European Conference on Computer Vision},
  784--799.

\bibitem[{Jocher(2020)}]{Jocher_YOLOv5_by_Ultralytics_2020}
Jocher, G. 2020.
\newblock {YOLOv5 by Ultralytics}.

\bibitem[{Li and Wu(2018)}]{li2018densefuse}
Li, H.; and Wu, X.-J. 2018.
\newblock DenseFuse: A fusion approach to infrared and visible images.
\newblock \emph{IEEE Transactions on Image Processing}, 28(5): 2614--2623.

\bibitem[{Li et~al.(2023{\natexlab{a}})Li, Xu, Wu, Lu, and
  Kittler}]{li2023lrrnet}
Li, H.; Xu, T.; Wu, X.-J.; Lu, J.; and Kittler, J. 2023{\natexlab{a}}.
\newblock Lrrnet: A novel representation learning guided fusion network for
  infrared and visible images.
\newblock \emph{IEEE transactions on Pattern Analysis and Machine
  Intelligence}, 45(9): 11040--11052.

\bibitem[{Li et~al.(2024)Li, Wang, Hu, Zhu, Li, and Wang}]{Li_2024_CVPR}
Li, K.; Wang, D.; Hu, Z.; Zhu, W.; Li, S.; and Wang, Q. 2024.
\newblock Unleashing Channel Potential: Space-Frequency Selection Convolution
  for SAR Object Detection.
\newblock In \emph{Proceedings of the IEEE/CVF Conference on Computer Vision
  and Pattern Recognition}, 17323--17332.

\bibitem[{Li et~al.(2023{\natexlab{b}})Li, Hou, Zheng, Cheng, Yang, and
  Li}]{li2023large}
Li, Y.; Hou, Q.; Zheng, Z.; Cheng, M.-M.; Yang, J.; and Li, X.
  2023{\natexlab{b}}.
\newblock Large selective kernel network for remote sensing object detection.
\newblock In \emph{Proceedings of the IEEE/CVF International Conference on
  Computer Vision}, 16794--16805.

\bibitem[{Lin et~al.(2017)Lin, Goyal, Girshick, He, and
  Doll{\'a}r}]{lin2017focal}
Lin, T.-Y.; Goyal, P.; Girshick, R.; He, K.; and Doll{\'a}r, P. 2017.
\newblock Focal loss for dense object detection.
\newblock In \emph{Proceedings of the IEEE/CVF International Conference on
  Computer Vision}, 2980--2988.

\bibitem[{Liu et~al.(2022{\natexlab{a}})Liu, Fan, Huang, Wu, Liu, Zhong, and
  Luo}]{liu2022target}
Liu, J.; Fan, X.; Huang, Z.; Wu, G.; Liu, R.; Zhong, W.; and Luo, Z.
  2022{\natexlab{a}}.
\newblock Target-aware dual adversarial learning and a multi-scenario
  multi-modality benchmark to fuse infrared and visible for object detection.
\newblock In \emph{Proceedings of the IEEE/CVF Conference on Computer Vision
  and Pattern Recognition}, 5802--5811.

\bibitem[{Liu et~al.(2023)Liu, Liu, Wu, Ma, Liu, Zhong, Luo, and
  Fan}]{liu2023multi}
Liu, J.; Liu, Z.; Wu, G.; Ma, L.; Liu, R.; Zhong, W.; Luo, Z.; and Fan, X.
  2023.
\newblock Multi-interactive feature learning and a full-time multi-modality
  benchmark for image fusion and segmentation.
\newblock In \emph{Proceedings of the IEEE/CVF International Conference on
  Computer Vision}, 8115--8124.

\bibitem[{Liu et~al.(2016)Liu, Anguelov, Erhan, Szegedy, Reed, Fu, and
  Berg}]{liu2016ssd}
Liu, W.; Anguelov, D.; Erhan, D.; Szegedy, C.; Reed, S.; Fu, C.-Y.; and Berg,
  A.~C. 2016.
\newblock Ssd: Single shot multibox detector.
\newblock In \emph{Proceedings of the European Conference on Computer Vision},
  21--37.

\bibitem[{Liu et~al.(2022{\natexlab{b}})Liu, Mao, Wu, Feichtenhofer, Darrell,
  and Xie}]{liu2022convnet}
Liu, Z.; Mao, H.; Wu, C.-Y.; Feichtenhofer, C.; Darrell, T.; and Xie, S.
  2022{\natexlab{b}}.
\newblock A convnet for the 2020s.
\newblock In \emph{Proceedings of the IEEE/CVF Conference on Computer Vision
  and Pattern Recognition}, 11976--11986.

\bibitem[{Nair and Hinton(2010)}]{nair2010rectified}
Nair, V.; and Hinton, G.~E. 2010.
\newblock Rectified linear units improve restricted boltzmann machines.
\newblock In \emph{Proceedings of the International Conference on Machine
  Learning}, 807--814.

\bibitem[{Razakarivony and Jurie(2016)}]{razakarivony2016vehicle}
Razakarivony, S.; and Jurie, F. 2016.
\newblock Vehicle detection in aerial imagery: A small target detection
  benchmark.
\newblock \emph{Journal of Visual Communication and Image Representation}, 34:
  187--203.

\bibitem[{Redmon et~al.(2016)Redmon, Divvala, Girshick, and
  Farhadi}]{redmon2016you}
Redmon, J.; Divvala, S.; Girshick, R.; and Farhadi, A. 2016.
\newblock You only look once: Unified, real-time object detection.
\newblock In \emph{Proceedings of the IEEE Conference on Computer Vision and
  Pattern Recognition}, 779--788.

\bibitem[{Ren et~al.(2015)Ren, He, Girshick, and Sun}]{ren2015faster}
Ren, S.; He, K.; Girshick, R.; and Sun, J. 2015.
\newblock Faster r-cnn: Towards real-time object detection with region proposal
  networks.
\newblock \emph{Advances in Neural Information Processing Systems}, 28.

\bibitem[{Robbins and Monro(1951)}]{robbins1951stochastic}
Robbins, H.; and Monro, S. 1951.
\newblock A stochastic approximation method.
\newblock \emph{The Annals of Mathematical Statistics}, 400--407.

\bibitem[{Singh, Najibi, and Davis(2018)}]{singh2018sniper}
Singh, B.; Najibi, M.; and Davis, L.~S. 2018.
\newblock Sniper: Efficient multi-scale training.
\newblock \emph{Advances in Neural Information Processing Systems}, 31.

\bibitem[{Xu et~al.(2020)Xu, Ma, Jiang, Guo, and Ling}]{xu2020u2fusion}
Xu, H.; Ma, J.; Jiang, J.; Guo, X.; and Ling, H. 2020.
\newblock U2Fusion: A unified unsupervised image fusion network.
\newblock \emph{IEEE Transactions on Pattern Analysis and Machine
  Intelligence}, 44(1): 502--518.

\bibitem[{Zhang et~al.(2020)Zhang, Fromont, Lefevre, and
  Avignon}]{zhang2020multispectral}
Zhang, H.; Fromont, E.; Lefevre, S.; and Avignon, B. 2020.
\newblock Multispectral fusion for object detection with cyclic fuse-and-refine
  blocks.
\newblock In \emph{2020 IEEE International Conference on Image Processing},
  276--280.

\bibitem[{Zhang and Ma(2021)}]{zhang2021sdnet}
Zhang, H.; and Ma, J. 2021.
\newblock SDNet: A versatile squeeze-and-decomposition network for real-time
  image fusion.
\newblock \emph{International Journal of Computer Vision}, 129(10): 2761--2785.

\bibitem[{Zhang et~al.(2023{\natexlab{a}})Zhang, Lei, Xie, Fang, Li, and
  Du}]{zhang2023superyolo}
Zhang, J.; Lei, J.; Xie, W.; Fang, Z.; Li, Y.; and Du, Q. 2023{\natexlab{a}}.
\newblock SuperYOLO: Super resolution assisted object detection in multimodal
  remote sensing imagery.
\newblock \emph{IEEE Transactions on Geoscience and Remote Sensing}, 61: 1--15.

\bibitem[{Zhang et~al.(2019)Zhang, Zhu, Chen, Yang, Lei, and
  Liu}]{Zhang_2019_ICCV}
Zhang, L.; Zhu, X.; Chen, X.; Yang, X.; Lei, Z.; and Liu, Z. 2019.
\newblock Weakly Aligned Cross-Modal Learning for Multispectral Pedestrian
  Detection.
\newblock In \emph{Proceedings of the IEEE/CVF International Conference on
  Computer Vision}.

\bibitem[{Zhang et~al.(2023{\natexlab{b}})Zhang, Li, Zhang, Zhang, Xu, Zhang,
  and Wang}]{10163477}
Zhang, R.; Li, L.; Zhang, Q.; Zhang, J.; Xu, L.; Zhang, B.; and Wang, B.
  2023{\natexlab{b}}.
\newblock Differential Feature Awareness Network within Antagonistic Learning
  for Infrared-Visible Object Detection.
\newblock \emph{IEEE Transactions on Circuits and Systems for Video
  Technology}, 1--1.

\bibitem[{Zhang et~al.(2023{\natexlab{c}})Zhang, Zhai, Liu, Wang, and
  Sun}]{zhang2023real}
Zhang, X.; Zhai, H.; Liu, J.; Wang, Z.; and Sun, H. 2023{\natexlab{c}}.
\newblock Real-time infrared and visible image fusion network using adaptive
  pixel weighting strategy.
\newblock \emph{Information Fusion}, 99: 101863.

\bibitem[{Zhang et~al.(2024)Zhang, Zhang, Wang, Ying, Sheng, Yu, Li, and
  Shen}]{zhang2024tfdet}
Zhang, X.; Zhang, X.; Wang, J.; Ying, J.; Sheng, Z.; Yu, H.; Li, C.; and Shen,
  H.-L. 2024.
\newblock Tfdet: Target-aware fusion for rgb-t pedestrian detection.
\newblock \emph{IEEE Transactions on Neural Networks and Learning Systems}.

\bibitem[{Zhang et~al.(2023{\natexlab{d}})Zhang, Yu, He, Wang, and
  Yang}]{zhang2023illumination}
Zhang, Y.; Yu, H.; He, Y.; Wang, X.; and Yang, W. 2023{\natexlab{d}}.
\newblock Illumination-guided RGBT object detection with inter-and
  intra-modality fusion.
\newblock \emph{IEEE Transactions on Instrumentation and Measurement}, 72:
  1--13.

\bibitem[{Zhao et~al.(2023{\natexlab{a}})Zhao, Xie, Zhao, He, and
  Lu}]{zhao2023metafusion}
Zhao, W.; Xie, S.; Zhao, F.; He, Y.; and Lu, H. 2023{\natexlab{a}}.
\newblock Metafusion: Infrared and visible image fusion via meta-feature
  embedding from object detection.
\newblock In \emph{Proceedings of the IEEE/CVF Conference on Computer Vision
  and Pattern Recognition}, 13955--13965.

\bibitem[{Zhao et~al.(2023{\natexlab{b}})Zhao, Bai, Zhang, Zhang, Xu, Lin,
  Timofte, and Van~Gool}]{zhao2023cddfuse}
Zhao, Z.; Bai, H.; Zhang, J.; Zhang, Y.; Xu, S.; Lin, Z.; Timofte, R.; and
  Van~Gool, L. 2023{\natexlab{b}}.
\newblock Cddfuse: Correlation-driven dual-branch feature decomposition for
  multi-modality image fusion.
\newblock In \emph{Proceedings of the IEEE/CVF Conference on Computer Vision
  and Pattern Recognition}, 5906--5916.

\bibitem[{Zhao et~al.(2023{\natexlab{c}})Zhao, Bai, Zhu, Zhang, Xu, Zhang,
  Zhang, Meng, Timofte, and Van~Gool}]{zhao2023ddfm}
Zhao, Z.; Bai, H.; Zhu, Y.; Zhang, J.; Xu, S.; Zhang, Y.; Zhang, K.; Meng, D.;
  Timofte, R.; and Van~Gool, L. 2023{\natexlab{c}}.
\newblock DDFM: denoising diffusion model for multi-modality image fusion.
\newblock In \emph{Proceedings of the IEEE/CVF International Conference on
  Computer Vision}, 8082--8093.

\bibitem[{Zhou, Chen, and Cao(2020)}]{zhou2020improving}
Zhou, K.; Chen, L.; and Cao, X. 2020.
\newblock Improving multispectral pedestrian detection by addressing modality
  imbalance problems.
\newblock In \emph{Proceedings of the European Conference on Computer Vision},
  787--803.

\end{thebibliography}

\end{document}